\pdfoutput=1
\documentclass[11pt]{article}

\usepackage[T1]{fontenc}
\usepackage[utf8]{inputenc}
\usepackage{mathptmx}
\usepackage[scaled=0.92]{helvet}
\usepackage{courier}
\usepackage[letterpaper,margin=1in]{geometry}
\usepackage{microtype}
\usepackage{amsmath,amssymb}
\usepackage{graphicx}
\usepackage{tikz}
\usetikzlibrary{arrows.meta,calc}
\usepackage{booktabs}
\usepackage{tabularx}
\usepackage{array}
\usepackage{needspace}
\usepackage{placeins}
\usepackage{caption}
\usepackage[round,authoryear]{natbib}
\usepackage{xcolor}
\usepackage{listings}
\usepackage{algorithm}
\usepackage{algpseudocode}
\usepackage{xurl}
\usepackage[colorlinks=true,linkcolor=blue!50!black,citecolor=blue!50!black,urlcolor=blue!50!black]{hyperref}

\newcolumntype{L}{>{\raggedright\arraybackslash}X}
\lstdefinestyle{json}{
  basicstyle=\ttfamily\footnotesize,
  backgroundcolor=\color{gray!8},
  frame=none,
  breaklines=true,
  breakatwhitespace=false,
  columns=fullflexible,
  keepspaces=true,
  showstringspaces=false,
  xleftmargin=1em,
  xrightmargin=1em,
  aboveskip=0.3em,
  belowskip=0.3em
}
\title{Contestable Agents for Data Governance: Recording Human Stewardship of AI-Generated Metadata Across Structured and Unstructured Systems}
\author{
  Humza Akhtar\textsuperscript{1} \qquad Brian Freeman\textsuperscript{2} \\[0.8em]
  \small \textsuperscript{1}Industry Solutions, MongoDB, Inc. \\
  \small \textsuperscript{2}Advanced Concepts \& Capabilities, Trane Technologies \\[0.5em]
  \small \texttt{humza.akhtar@mongodb.com}, \texttt{brian.freeman@tranetechnologies.com}
}

\date{}

\begin{document}
\maketitle

\begin{abstract}
Industrial organizations hold operational knowledge in structured records, such as telemetry, point lists, and transactions, and in unstructured sources, such as manuals, service notes, communications, and inspection reports. Governance programs often manage these sources separately. Classifying, linking, and documenting them depends on a small group of employees with deep domain knowledge. Agents built on large language models can propose metadata, cite documents, and flag quality issues, but unreviewed proposals can spread errors into downstream systems. A perfunctory approval leaves no record of what a reviewer checked. We present contestable governance: specialist agents produce scored, cited arguments for disputed metadata decisions; a steward can accept, reject, modify, or add arguments before commit; and a governance record retains the evidence, interventions, and outcome. A versioned domain glossary supplies model-specific terms and acronym definitions to the agents. We describe the workflow, a triage path that reserves full argumentation for uncertain proposals while preserving review before commit, the scoring and record models, and a gate that controls which verified decisions become precedent. Two HVAC use cases illustrate the pattern: point classification using Brick Schema class labels, supported by an open reference implementation, and an illustrative warranty failure-mode decision linking claim fields and telemetry to service notes and revisioned bulletins.

\end{abstract}

\section{Introduction}

Manufacturers of commercial HVAC equipment depend on data in two forms. Connected chillers, rooftop units, and air handlers stream structured telemetry from hundreds of points per unit, while installation, operation, and maintenance manuals define those points in prose that varies by model and firmware revision. Factory end-of-line tests produce structured pass and fail results, while PDF specifications and change notices explain the test limits and engineering deviations. Warranty and service systems hold claim codes and part numbers, while technicians describe failures in free-text notes. Governance programs typically assign these sources to different tools and owners, leaving employees to remember the links between fields and the documents that define them.

That arrangement breaks down as the installed base grows. Every new product line, firmware update, and connected site generates data that must be classified and linked to documentation before remote monitoring and service teams can use it reliably. Few employees have the equipment knowledge to do this work. Until a steward reviews a new model, its points may remain unclassified or carry incorrect classes. If a mixed-air temperature sensor is labeled as discharge air, for example, a remote monitoring system could interpret normal mixing as a failed heating coil and send a technician to the site with an incorrect diagnosis.

AI agents built on large language models change the economics of this work. Foundation models can match schemas, clean entries, and infer the meaning of cryptic field names with little task-specific training \citep{narayan2022}. Agents that combine a model with retrieval and extraction tools can read a manual, compare a field with prior decisions, and propose a classification \citep{wang2024,lewis2020}. They also introduce a governance problem. An agent that writes metadata directly to a catalog can propagate an error to every report and application that consumes it. A wrong class assigned to one air handler can spread across a portfolio when its mapping is copied to other units of the same model. Reviewers who approve output with a single click risk rubber-stamping it, a documented pattern in which people accept automated recommendations with little scrutiny \citep{parasuraman2010,bucinca2021}. The resulting record says that a person approved the decision without showing what they checked. When a reviewer overrules an agent, for example by reclassifying a supplier's part number, the correction may stay in an email or ticket that the agent never sees. The agent then repeats the mistake. Months later, an engineer investigating a fault rule that fires on the wrong point cannot tell whether its classification came from a model, a reviewer, or a bulk import.

Closing this gap requires seven properties, which we use as requirements throughout the paper: proposals must cite the structured fields and documents they rely on (R1); a reviewer must be able to contest them before commit (R2); every decision must be attributable to an agent, a model version, and a named reviewer, as applicable (R3); history must be append-only (R4); the full lineage of a decision must be retrievable in one query (R5); verified decisions must pass a gate before becoming precedent (R6); and reviewer interventions must be stored as measurable data (R7).

Here, \emph{lineage} means the trace from source evidence through agent proposals and human edits to committed metadata. We use PROV, the World Wide Web Consortium's provenance data model, to describe that trace: its \emph{entities} include documents and decision records, its \emph{activities} include generating and reviewing arguments, and its \emph{agents} are the models and people responsible \citep{moreau2013}. PROV provides a conceptual mapping in this paper; the reference implementation does not store records in a PROV serialization.

This paper argues that contestability belongs in the data layer of agent-assisted governance. A contestable system lets a qualified person challenge the reasoning before an output is committed and retains that intervention in the decision record \citep{hirsch2017,alfrink2023}. In our proposed pattern, agents attach cited evidence to each decision, specialist agents argue disputed cases, and a steward can contest a proposal before commit. A governance record shows who made each decision and how the steward intervened (Section~\ref{sec:approach}). We discuss two HVAC use cases (Section~\ref{sec:usecases}). The first is point classification, supported by an open reference implementation. The second is an illustrative warranty failure-mode classification that shows how a steward could resolve conflicting evidence across structured service records and revisioned documents.

\section{Background and Related Work}

\citet{khatri2010} frame data governance as a set of decision domains covering data principles, data quality, metadata, data access, and data lifecycle, with explicit assignment of who holds decision rights in each. \citet{weber2009} show that the right governance structure depends on organizational context, and \citet{otto2011} documents how large firms organize stewardship roles in practice. \citet{abraham2019} consolidate this literature into a framework of structural, procedural, and relational mechanisms, and the DAMA body of knowledge sets out the operational practices stewards follow \citep{dama2017}. Across this work, stewardship is a human activity performed alongside operations, and its outputs, such as approved definitions and classifications, are stored separately from the reasoning that produced them.

Large data catalogs show what happens when documentation depends entirely on engineers: Google's Goods system infers metadata for billions of datasets after creation because requiring documentation up front did not scale \citep{halevy2016}. Provenance research distinguishes why a result was produced from where its data came from \citep{buneman2001}. PROV formalizes this account through relationships among entities, activities, and agents \citep{moreau2013}. Other work describes what metadata should enable. The FAIR principles call for data to be findable, accessible, interoperable, and reusable \citep{wilkinson2016}, while datasheets ask producers to document how datasets were collected and processed \citep{gebru2021}. Data quality provides a way to judge the result across dimensions that depend on fitness for use \citep{wang1996}. Together, these works motivate recording evidence and responsibility close to the decision that creates metadata.

Point naming in building automation makes the metadata problem concrete. A controller exposes thousands of points, and one vendor may label an air handler's supply-air temperature sensor \texttt{AHU3\_SA\_T}, while another uses \texttt{SupplyAirTemp\_03}. Brick defines a shared ontology of building equipment, points, and their relationships \citep{balaji2016}. Vendors still name points differently, so someone must map raw names to Brick classes before energy reports and fault detection rules can use them consistently. Automated approaches infer mappings from names, for example by learning parsing rules from a few expert examples \citep{bhattacharya2015} or transferring learned mappings across buildings through an intermediate representation \citep{koh2018}. A name alone can mislead when vendor conventions differ, and relevant manuals may resolve the ambiguity. Foundation models can read those manuals \citep{narayan2022}; the question is how to use both kinds of evidence while keeping a reviewer in control of what is committed.

\citet{freeman2026epistemic} compared four sequential single-task agents with one multi-task agent on a fixed ERP incident, then added a reconciler to inspect contradictions. They also tested glossary injection on a separate HVAC troubleshooting query. Their judge-based results motivate role separation and explicit vocabulary here, but do not establish that either measure improves the factual correctness of metadata decisions in this workflow.

Contestability has been studied as a design goal for machine learning systems whose users need to question outputs as part of their professional practice \citep{hirsch2017}. Later work examined how people understand challenges to algorithmic decisions \citep{lyons2021} and proposed design practices for contestable AI \citep{alfrink2023}. These studies describe what a person should be able to do; computational argumentation offers one way to represent what they can challenge. Abstract argumentation frameworks model arguments and their attacks on one another \citep{dung1995}, while bipolar frameworks add support relations \citep{cayrol2005}. Quantitative bipolar argumentation frameworks (QBAFs) assign base scores and compute how support and attack affect argument strength, as in DF-QuAD \citep{rago2016}. An argument graph divides a decision into claims that a person can challenge individually. Such graphs have been proposed for contestable AI \citep{leofante2024}, and recent work constructs them from language-model-generated arguments \citep{freedman2024}. We apply this approach to data governance and add a storage model that preserves each reviewer's edits alongside the arguments they changed.

\section{Our Approach}
\label{sec:approach}

\subsection{Workflow}
The full contestation path has four phases. In \textbf{data analysis}, an analyzer reads structured inputs, such as a point list or schema, gathers relevant documents and glossary entries, and retrieves similar verified decisions from a precedent library. In \textbf{multi-agent argumentation}, specialist agents submit supporting or challenging arguments only when their evidence bears on a candidate; each argument has a base score and a source reference (R1). Lack of evidence for one candidate does not, by itself, support another. The agents have distinct evidence roles: one might inspect a point name, another the equipment manual, and a third the applicability of a document revision. Their disagreements show a reviewer where the evidence diverges. In \textbf{human contestation}, a steward can accept, reject, modify, or add arguments. The system rescores after each change, and no decision is committed before review ends (R2). In \textbf{selection and storage}, the scoring engine selects the highest-scoring candidate, a finalizer explains the outcome and propagation steps, and the system stores the evidence, arguments, interventions, and decision in a governance record (R3 to R5). The finalizer does not choose the class; an auditor can recompute it from the stored arguments without rerunning a model. Algorithm~\ref{alg:contestation-workflow} summarizes this proposed sequence.

\begin{algorithm}
\caption{Simplified Contestation Workflow with Human Oversight}
\label{alg:contestation-workflow}
\begin{algorithmic}[1]
\Procedure{Contest}{$input$}
    \Statex \textbf{Phase 1: Data analysis}
    \State $candidates \gets$ AnalyzerIdentifyCandidates($input$)
    \State $evidence \gets$ AnalyzerGatherDocumentsGlossaryAndPrecedents($input$)

    \Statex \textbf{Phase 2: Multi-agent argumentation}
    \State $arguments \gets$ AgentsCollectRelevantArguments($candidates$, $evidence$)
    \State $scores \gets$ ScoringEngineCalculateScores($candidates$, $arguments$)

    \Statex \textbf{Phase 3: Human contestation}
    \State $stewardActions \gets []$
    \While{review is open}
        \State $action \gets$ HumanStewardAcceptRejectModifyOrAddArgument()
        \State Apply($action$)
        \State Append($stewardActions$, $action$)
        \State $scores \gets$ ScoringEngineRecalculateScores($candidates$, $arguments$)
    \EndWhile

    \Statex \textbf{Phase 4: Selection, explanation, and storage}
    \State $selected \gets$ ScoringEngineSelectHighestScoringCandidate($scores$)
    \State $explanation \gets$ FinalizerExplainResultAndPropagation($selected$, $arguments$, $scores$)
    \State $record \gets$ StoreGovernanceRecord($input$, $evidence$, $arguments$,
    \Statex \hspace{1.5em}$stewardActions$, $selected$, $explanation$)
    \State \Return $record$
\EndProcedure

\Procedure{Audit}{$record$}
    \State $recomputed \gets$ RecomputeClassFromStoredArguments($record.arguments$)
    \State \Return Compare($recomputed$, $record.selected$)
\EndProcedure
\end{algorithmic}
\end{algorithm}

\paragraph{Role constraints.}
The human steward may accept, reject, modify, or add arguments during review. The scoring engine selects the highest-scoring candidate. The finalizer explains the result and propagation steps but does not choose the class. The auditor can recompute the class from the stored arguments without rerunning a model.

\subsection{Domain vocabulary}
\label{sec:glossary}
Acronyms and equipment terms often change meaning across vendors, product families, and revisions. Before the agents evaluate a case, the analyzer can supply a controlled glossary containing only terms present in its evidence packet. Each entry needs an expansion, a definition, its equipment and revision scope, the source passage that authorizes it, a version, and a steward owner. For example, \texttt{DISCH} may identify the discharge side of a mixing box on one air handler but a post-coil location on another. A universal expansion would conceal that difference. Selecting relevant entries also avoids filling each agent's context with unrelated definitions, a tradeoff explored in the glossary experiments of \citet{freeman2026epistemic}.

The glossary is governed metadata, not an unquestionable prompt instruction. Agents should cite both an applicable glossary entry and its source document when a term affects a proposal. A reviewer who finds an obsolete or ambiguous definition can challenge it, record the correction, and prevent the unverified entry from becoming precedent. The proposed decision record would store the glossary version used so that later changes can identify decisions needing review.

\subsection{Triage}
\label{sec:triage}
Running every decision through full argumentation requires model calls and reviewer time, even though many points on a new unit are unambiguous. A typed decision model such as Jev, which returns a bounded choice with a probability instead of generated text \citep{typesafe2026}, could classify points first and send only low-confidence proposals through multi-agent argumentation. In a separate controlled benchmark of 100 agent tasks, a similar routing design reduced calls to a strong language model by 72.7\% while reaching 95\% task success \citep{wu2026reflex}. Those figures do not measure this governance workflow.

High-confidence proposals still go to a steward with their source evidence, model version, and probability. Each remains contestable before an explicit review decision and catalog commit (R2). Its governance record stores the question, evaluated state, reviewer action, and outcome even when no argument graph was generated (R3). The threshold that selects the review path is itself a governance decision: it needs a named owner and must be stored with every record it affects. A reviewer can escalate any proposal to full argumentation because a probability alone gives no specific reasoning to challenge.

\subsection{Argument scoring}
\label{sec:scoring}
Each decision on the full contestation path uses a quantitative bipolar argumentation framework (QBAF) $\langle A, R^{+}, R^{-}, \tau \rangle$, drawing on bipolar and quantitative argumentation \citep{cayrol2005,rago2016}. ``Bipolar'' means that arguments can support or attack one another; ``quantitative'' means that they have numerical strengths. The set $A$ contains the active arguments. For an argument $x$, $R^{+}(x)$ contains its supporters, $R^{-}(x)$ contains its attackers, and $\tau(x)\in[0,1]$ is its unitless base score. The framework specifies the range of $\tau(x)$ but does not determine its value: an assessor must assign that input before the graph is scored. Within a candidate, supporting arguments support one another, as do challenging arguments. Every supporting argument attacks every challenging argument, and vice versa. Let $f^{t}(x)$ denote the computed strength of argument $x$ after iteration $t$, with $f^{0}(x)=\tau(x)$.

We use the signed-sum logistic update from the open reference implementation as an illustrative QBAF scoring rule. Its support and attack contributions have explicit signs, its inputs can be stored for independent recomputation, and the logistic function bounds strengths in $(0,1)$. This choice preserves the reference implementation's arithmetic for the worked examples; it is not evidence that this rule is preferable. DF-QuAD \citep{rago2016} and other modular weighted semantics for bipolar graphs \citep{potyka2019} offer alternative ways to aggregate and propagate support and attack. The unitless coefficient $\alpha$ scales every support and attack contribution before the logistic transform. We fix $\alpha=0.3$ as an illustrative uniform edge weight, not a parameter fitted to outcomes. These strengths are comparison scores, not calibrated probabilities. For a positive integer $n$, we update all strengths simultaneously for $n$ iterations:
\begin{equation}
  f^{t+1}(x) = \operatorname{sig}\!\Big( \tau(x) + \alpha \sum_{y \in R^{+}(x)} f^{t}(y) \;-\; \alpha \sum_{z \in R^{-}(x)} f^{t}(z) \Big), \qquad t=0,\ldots,n-1.
  \label{eq:update}
\end{equation}
\Needspace{4\baselineskip}
Here, $\operatorname{sig}(u)=1/(1+e^{-u})$ is the logistic function. The summation variable $y$ ranges over arguments supporting $x$, and $z$ ranges over arguments attacking $x$; $f^{t}(y)$ and $f^{t}(z)$ are their strengths at iteration $t$. The value $f^{n}(x)$ is the computed strength of argument $x$ after $n$ updates. A candidate's raw margin is the sum of its supporting arguments' final strengths minus the sum of its challenging arguments' final strengths.

The first update uses each neighbor's base score $f^{0}=\tau$, before that neighbor's own supporters and attackers have changed its strength. Later updates propagate those changes through the graph, including its reciprocal links, until the strengths approach a fixed point. Iteration seeks a self-consistent graph score; it does not add evidence or change $\tau$.

\Needspace{10\baselineskip}
For the worked warranty example, we set $n=10$, matching the reference implementation's default\footnote{\url{https://github.com/mongodb-industry-solutions/contestable-ai/blob/main/argumentation/qbaf.py}} and serving as a case-specific numerical stopping count. S1 moves from $f^{1}=0.665$ to $f^{10}=0.683$. In the largest candidate graph of that example (Section~\ref{sec:coauthor}), each of five arguments has at most four incoming relations. Since the logistic derivative is at most $1/4$, Equation~\ref{eq:update} has a sup-norm contraction bound $q\leq(1/4)(0.3)(4)=0.3$. The saved reproducibility script finds a maximum $\max_x|f^{10}(x)-f^{9}(x)|$ of $3.20\times10^{-7}$ across the candidate graphs before and after review. The contraction-mapping error bound \citep{rudin1976analysis} multiplies this difference by $q/(1-q)$, making the remaining error at most $1.37\times10^{-7}$ per argument and $6.85\times10^{-7}$ per raw margin, below the three-decimal precision used to compare candidates. A local sensitivity check at $\alpha=0.2$ and $0.4$ preserves Sensor as the pre-review winner and Linkage as the post-review winner, although the margins change. These checks establish numerical stability for this example, not decision validity, calibration, or general robustness; larger graphs or different weights require a new convergence check.

The reference implementation normalizes candidate margins for display; the worked examples report raw margins. Reviewer operations change the active graph: a rejected argument is removed before rescoring but retained in the record; an added argument enters with a base score of 0.85; a modified argument retains its original text; and an accepted argument is marked as reviewed. With no incoming support or attack, the logistic transform maps base scores in $[0,1]$ to strengths between 0.50 and 0.73, rather than preserving an isolated argument's base score. This behavior should be assessed against published axioms for weighted bipolar argumentation \citep{amgoud2018weighted}. The presence or absence of an argument can therefore move a candidate's score more than a modest change in its base score, making rejection a strong review action. In the reference implementation, the agent that writes an argument also assigns its base score. For the synthetic warranty example, we selected the base scores in Table~\ref{tab:warranty_arguments} to illustrate the scoring and review steps; they were not derived from a rubric, empirical calibration, or a rating study. In a deployment, a versioned rubric would first check whether the cited source applies to the item and revision: an inapplicable source should not contribute a scored argument. For each eligible argument, the rubric could assess the directness and specificity of the evidence and record the assessor, rationale, and rubric version. Independent adjudication would then be needed to test whether these assignments are repeatable and improve decisions. A separate decision model could score arguments, separating authorship from assessment.

\subsection{Governance record}
\label{sec:record}
The governance record connects a decision with its evidence and review history, so an auditor can retrieve its lineage without manually reconciling logs, tickets, and catalogs. The reference implementation uses MongoDB documents for arguments and audit entries, time series collections for sensor data, and vector search over embedded documents. In the proposed cross-system schema, evidence sits in a context document; each argument has its own document with base and final scores and edit fields; and the decision document contains the outcome, scores before and after review, and an entry for every human edit. A shared session identifier connects these documents, allowing one aggregation pipeline to return the decision's lineage (R5):

\Needspace{9\baselineskip}
\begin{lstlisting}
db.governance_records.aggregate([
  { $match: { session_id: "a7c3..." } },
  { $lookup: { from: "argument_records", localField: "session_id",
               foreignField: "session_id", as: "arguments" } },
  { $lookup: { from: "session_contexts", localField: "session_id",
               foreignField: "session_id", as: "evidence" } },
  { $unwind: "$evidence" }
])
\end{lstlisting}

This structure can be expressed in PROV terms \citep{moreau2013}: source documents and arguments are entities; generation, review, and edits are activities; and agents and reviewers are responsible actors. The decision record is the resulting entity. Storing interventions as data also supports operational questions: which agent's arguments are rejected most often, how long review takes for a new equipment model, and how many points the triage threshold routes to full argumentation (R7). To meet the append-only requirement in a deployment, edits would need a sequenced event log and database roles that allow inserts but prohibit updates and deletes (R4). Schema validation should require a named reviewer, model and prompt versions, and a rationale for each intervention (R3).

\subsection{Promotion gate}
\label{sec:loop}
A verified record is useful both as a decision and as precedent for a similar case. Its embedding could enter the vector index searched during data analysis, allowing the next unit of the same equipment model to retrieve the reviewer's corrections. An incorrect precedent could spread across a portfolio. The proposed promotion gate therefore admits only verified records (R6), stores each promotion as an attributable event, and lets a reviewer demote a record. Demotion would flag decisions that relied on the record for review.

\section{Use Cases}
\label{sec:usecases}

\subsection{Use case 1: HVAC point classification}
\label{sec:hvac}

\begin{figure}[ht]
\centering
\includegraphics[width=\textwidth]{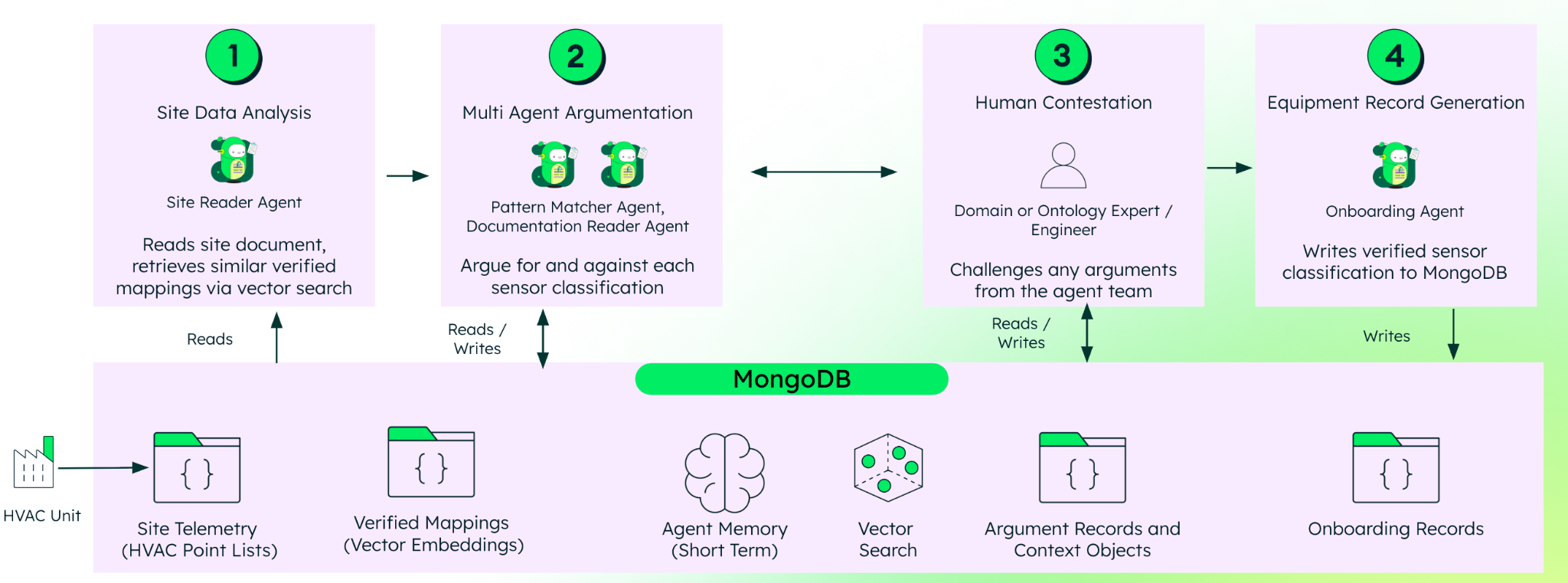}
\caption{Contestable governance for HVAC point classification: data analysis, multi-agent argumentation between two agents, human contestation, and selection and storage of a verified onboarding record.}
\label{fig:hvac}
\end{figure}

Each connected HVAC unit exposes hundreds of points with vendor-specific names. Mapping them to standard classes lets building management systems and analytics platforms interpret them consistently. The open reference implementation \url{https://github.com/mongodb-industry-solutions/iomt-contestable-ai} demonstrates contestable governance for this task on synthetic data using CrewAI agents, GPT-4o, Voyage AI embeddings, and MongoDB (Figure~\ref{fig:hvac}).

The synthetic site contains 247 points, 18 of them flagged as ambiguous. Under the proposed triage path, 229 points would receive a shorter evidence-based steward review, while the 18 ambiguous points would receive full argumentation and review. None would be committed before review. The reference implementation runs the full workflow on one point: \texttt{AHU-1-DISCH-T}, on an air handler from a fictional EQ-7700 series. The analyzer retrieves precedents showing why this point needs review: \texttt{DISCH-T} maps to mixed air on EQ-7700 air handlers and supply air on EQ-7700 fan coil units, while a similar name on another vendor's equipment refers to post-coil discharge air. Two agents argue over three option labels: \nolinkurl{brick:Supply_Air_Temperature_Sensor}, \nolinkurl{brick:Discharge_Air_Temperature_Sensor}, and \nolinkurl{brick:Mixed_Air_Temperature_Sensor}. Brick 1.4.2 declares the supply-air and discharge-air sensor classes equivalent and marks the latter as an alias of the former \citep{brick142}. The three labels therefore represent two distinct ontology choices; the reference implementation nevertheless scores the labels separately. The String Matcher sees only the point name, the evidence available to name-based methods \citep{bhattacharya2015,koh2018}, and favors discharge air. The Documentation Reader sees manual fragments and cites section 4.2, which places the sensor in the mixed-air plenum, and Appendix~A, which explains that \texttt{DISCH} refers to the discharge side of the mixing box.

A model-scoped glossary would preserve these different meanings of \texttt{DISCH} with the applicable equipment type and manual revision. A single global expansion would conceal the conflict that the reviewer needs to inspect.

With illustrative base scores, the two leading candidates finish argumentation nearly tied: a margin of 0.045 for discharge air and 0.044 for mixed air. The reviewer rejects the String Matcher's challenge to mixed air because it rests on the abbreviation alone, raising the mixed-air margin to 0.821. The reviewer then adds an argument citing section 4.2, raising the margin to 1.758. The system commits mixed air, consistent with the documentation. The record attributes the class to two named agents and a reviewer, cites the decisive manual section, and preserves the rejected reasoning and the reason for rejection. If promoted, the decision would give the next EQ-7700 air handler a verified precedent that includes the reviewer's knowledge of this vendor's use of \texttt{DISCH}.

The scenario also exposes decisions a production deployment would need to make. Normalized scores display the near tie above as 1.00 and 0.99, which a reviewer might mistake for high confidence; raw margins should appear alongside them. Base scores are assigned by the model itself, remain uncalibrated, and can vary between runs. The reference implementation overwrites an argument on a second edit, seeds its precedent library without writing verified decisions back, and has no triage step. Append-only history, the promotion gate, and triage remain proposed extensions.

\subsection{Use case 2: Governing warranty failure-mode metadata across systems}
\label{sec:coauthor}

A fictional air handler, model \texttt{DEMO-AHU} with serial \texttt{DEMO-2407}, has a claim record C1 containing a supply air temperature (\texttt{SAT}) fault code and a replaced supply-air sensor. A telemetry window T1 shows a changing damper command while measured position stays fixed. A technician note span N1 reports a loose damper linkage and says the sensor replacement did not clear the fault. A service bulletin, B1, describes sensor drift only for serial suffixes 1000--1999. Glossary entry G1 records this model-specific expansion of \texttt{SAT}. The governed field is one failure-mode tag with two candidates: \emph{sensor drift} (Sensor) and \emph{damper linkage failure} (Linkage). A claim-to-bulletin link is a separate decision. The original claim code remains source data.

\begin{figure}[!ht]
\centering
\definecolor{wfink}{HTML}{123B3B}
\definecolor{wfgreen}{HTML}{00D56F}
\definecolor{wflavender}{HTML}{FBF2FF}
\definecolor{wfpalegreen}{HTML}{EFFDF5}
\begin{tikzpicture}[
  x=1cm,y=1cm,
  font=\sffamily,
  wfstage/.style={draw=wfink!38, line width=0.55pt, rounded corners=3pt,
    fill=wflavender, minimum width=10.50cm},
  wfresult/.style={wfstage, fill=wfpalegreen},
  wfagent/.style={draw=wfink!38, line width=0.45pt, rounded corners=2pt,
    fill=white, minimum width=2.80cm, minimum height=1.85cm},
  wfbadge/.style={circle, draw=wfink, line width=0.6pt, fill=wfgreen,
    minimum size=0.52cm, inner sep=0pt, font=\bfseries\small, text=wfink},
  wftitle/.style={font=\bfseries\small, text=wfink, align=center},
  wfbody/.style={font=\footnotesize, text=wfink, align=center},
  wfarr/.style={-{Latex[length=2.4mm,width=1.5mm]}, line width=0.85pt, draw=wfink}
]
  \node[wfstage, minimum height=1.55cm] (packet) at (0,10.95) {};
  \node[wfstage, minimum height=3.75cm] (agents) at (0,8.00) {};
  \node[wfstage, minimum height=1.65cm] (review) at (0,4.80) {};
  \node[wfresult, minimum height=1.65cm] (record) at (0,2.50) {};
  \node[wfresult, minimum height=0.72cm] (store) at (0,0.55) {};

  \node[wfbadge] at ($(packet.north west)+(0.47,-0.38)$) {1};
  \node[wfbadge] at ($(agents.north west)+(0.47,-0.38)$) {2};
  \node[wfbadge] at ($(review.north west)+(0.47,-0.38)$) {3};
  \node[wfbadge] at ($(record.north west)+(0.47,-0.38)$) {4};

  \node[wftitle] at ($(packet.north)+(0,-0.32)$) {Versioned evidence packet};
  \node[wfbody] at ($(packet.center)+(0,-0.23)$)
    {Claim C1 and telemetry T1 \quad|\quad note N1 and bulletin B1\\Scoped glossary G1 defines \texttt{SAT} for the fictional model};

  \node[wftitle] at ($(agents.north)+(0,-0.34)$) {Specialist agents and cited arguments};
  \node[wfagent] at (-3.14,8.05) {};
  \node[wfagent] at (0,8.05) {};
  \node[wfagent] at (3.14,8.05) {};
  \foreach \x in {-3.14,0,3.14} {
    \draw[draw=wfink,fill=wfgreen!40,rounded corners=2pt]
      ({\x-0.28},8.43) rectangle ({\x+0.28},8.82);
    \fill[wfink] ({\x-0.11},8.62) circle (0.026);
    \fill[wfink] ({\x+0.11},8.62) circle (0.026);
    \draw[draw=wfink,line width=0.45pt] (\x,8.82) -- (\x,8.95);
    \fill[wfink] (\x,8.97) circle (0.035);
  }
  \node[wfbody, text width=2.55cm] at (-3.14,7.62)
    {Claims and telemetry\\C1, T1, G1};
  \node[wfbody, text width=2.55cm] at (0,7.62)
    {Service note\\N1, G1};
  \node[wfbody, text width=2.55cm] at (3.14,7.62)
    {Bulletin scope\\B1, C1};
  \node[wfbody] at (0,6.48)
    {Reconciler flags the S3/S5 conflict; agents submit cited support and challenge.};

  \node[wftitle] at ($(review.north)+(0,-0.32)$) {Human steward contests the arguments};
  \node[wfbody, text width=8.7cm] at ($(review.center)+(0,-0.24)$)
    {Reject out-of-scope S3; add S6 from N1.\quad Rescore both tags and authorize before commit.};

  \node[wftitle] at ($(record.north)+(0,-0.32)$) {Decision and governance record};
  \node[wfbody, text width=8.7cm] at ($(record.center)+(0,-0.24)$)
    {Tag: \emph{damper linkage failure}\\Evidence, agent arguments, and steward edits\\Separate bulletin link; verified precedent gate};

  \node[font=\scriptsize, text=wfink, align=center, text width=10.0cm] at (store.center)
    {Proposed store: source revisions \quad|\quad arguments \quad|\quad interventions \quad|\quad outcome};

  \draw[wfarr] (packet.south) -- (agents.north);
  \draw[wfarr] (agents.south) -- (review.north);
  \draw[wfarr] (review.south) -- (record.north);
  \draw[wfarr] (record.south) -- (store.north);
\end{tikzpicture}
\caption{Proposed workflow for the fully synthetic warranty example. Versioned evidence and glossary entries feed specialist agents; a steward contests the arguments before the selected tag and its lineage are stored. The promotion gate is a proposed extension.}
\label{fig:warranty}
\end{figure}

The analyzer snapshots C1, T1, N1, B1, and G1 in Figure~\ref{fig:warranty}, capturing the state of all relevant evidence at the time of analysis. A claims-and-telemetry agent receives C1, T1, and G1; a service-note agent receives N1 and G1; and a bulletin-scope agent receives B1 and C1's serial number. These agents submit arguments with cited evidence; a reconciler flags contradictions but cannot select a tag or authorize a commit. Table~\ref{tab:warranty_arguments} lists each candidate's supporting and challenging arguments with its cited basis and illustrative base score. The author-selected scores create a contested decision, rather than estimate evidential reliability. In this example, the bulletin title produces a provisional Sensor argument (S3) with $\tau=0.80$, while the serial-range check challenges it (S5). Because B1 does not apply to this unit, S3 is an erroneous argument to remove during review, not an applicable argument that merely deserves a lower score.

\begin{table}[htbp]
\centering
\caption{Synthetic arguments for the failure-mode decision. A plus sign supports a candidate and a minus sign challenges it. S3 is rejected and S6 is added during review. The author-selected $\tau$ values are illustrative inputs, not probabilities or the computed $f^{10}$ strengths.}
\label{tab:warranty_arguments}
\footnotesize
\begin{tabularx}{\textwidth}{@{}llLr@{}}
\toprule
ID & Effect & Cited basis & $\tau$ \\
\midrule
S1 & Sensor $+$ & C1: \texttt{SAT} fault code & 0.70 \\
S2 & Sensor $+$ & C1: supply-air sensor replaced & 0.65 \\
S3 & Sensor $+$ & B1: bulletin title mentions sensor drift & 0.80 \\
S4 & Sensor $-$ & T1: damper command changes; position stays fixed & 0.75 \\
S5 & Sensor $-$ & B1 and C1: serial ranges do not overlap & 0.75 \\
D1 & Linkage $+$ & T1: damper command-position mismatch & 0.80 \\
D2 & Linkage $+$ & N1: technician observed a loose linkage & 0.85 \\
D3 & Linkage $-$ & C1: fault code points toward a sensor & 0.70 \\
\midrule
S6 & Sensor $-$ & N1: sensor swap did not resolve fault; steward-added & 0.85 \\
\bottomrule
\end{tabularx}
\end{table}

\Needspace{13\baselineskip}
For a worked calculation, consider S1 before review. At $t=0$, its supporters S2 and S3 have base scores 0.65 and 0.80, while its attackers S4 and S5 each have base score 0.75. The first synchronous update of Equation~\ref{eq:update} is
\[
\begin{aligned}
f^{1}(\mathrm{S1})
  &=\operatorname{sig}\!\bigl(0.70+0.3(0.65+0.80)-0.3(0.75+0.75)\bigr)\\
  &=\operatorname{sig}(0.685)=0.665.
\end{aligned}
\]
After ten updates, $f^{10}(\mathrm{S1})=0.683$. The other Sensor strengths before review are 0.673 and 0.703 for S2 and S3, and 0.576 each for S4 and S5. Their signed sum gives a raw Sensor margin of $0.683+0.673+0.703-0.576-0.576=0.907$; the corresponding Linkage margin is 0.839. The arXiv ancillary files include the script \texttt{reproduce\_warranty\_scores.py}, which recomputes the per-argument $f^{1}$ and $f^{10}$ values and candidate margins, and its output, \texttt{warranty\_score\_results.txt}.

The reconciler flags the conflict between S3 and S5. The steward rejects S3 because B1 excludes serial \texttt{DEMO-2407} and adds S6 from N1. Rescoring changes $f^{10}(\mathrm{S1})$ to 0.557 and the Sensor margin to $-1.016$; the Linkage margin remains $0.839$. The scoring engine selects Linkage, and the steward checks the result before authorizing the commit. These numbers show how the defined scoring rule responds to edits; they are neither calibrated confidence values nor measured classification accuracy. Table~\ref{tab:warranty_language} contrasts the ambiguous shorthand in the synthetic records with the evidence-linked language available after glossary and source checks.

\begin{table}[htbp]
\centering
\caption{Synthetic examples of ambiguous language and clear, evidence-linked results. The right column states what the records support after glossary and source checks.}
\label{tab:warranty_language}
\footnotesize
\begin{tabularx}{\textwidth}{@{}p{0.29\textwidth}L@{}}
\toprule
Ambiguous shorthand & Clear, reviewable result \\
\midrule
\emph{SAT fault; changed sensor.} & G1 expands \texttt{SAT} as supply-air temperature. C1 records a code and part replacement; neither proves sensor drift. \\
\emph{Bulletin confirms drift.} & B1 revision B covers serial suffixes 1000--1999. C1 has suffix 2407, so this bulletin does not apply to the unit. \\
\emph{Damper seems bad.} & T1 shows command-position mismatch, and N1 records a loose linkage and unsuccessful sensor replacement. Proposed tag: \emph{damper linkage failure}. \\
\bottomrule
\end{tabularx}
\end{table}

\FloatBarrier

The proposed governance record retains versions of C1 through G1, the original claim code, both candidate tags, every argument, the S3 rejection and S6 addition, the reviewer identity, and the final tag. The separate claim-to-bulletin link would be rejected on applicability grounds. Only a verified record could become precedent; a later bulletin revision could flag dependent decisions for review.

The following abbreviated JSON documents illustrate how this synthetic decision and its S1 argument could be stored. The decision contains the selected tag, candidate margins, and steward edits; the argument contains the base score $\tau$ and computed $f^{10}$ strengths. Both share a \texttt{session\_id} with the evidence snapshot, allowing retrieval through the aggregation in Section~\ref{sec:record}.

\Needspace{34\baselineskip}
\noindent\textbf{Decision document (\texttt{governance\_records}).}
\begin{lstlisting}
{
  "_id": "decision-DEMO-2407",
  "session_id": "session-DEMO-2407",
  "decision_type": "failure_mode_tag",
  "subject": {"model": "DEMO-AHU", "serial": "DEMO-2407"},
  "evidence_snapshot_id": "context-DEMO-2407",
  "glossary_version": "G1-v1",
  "scoring": {"rule": "logistic_qbaf_v1", "alpha": 0.3,
              "iterations": 10},
  "raw_margins": {
    "before": {"Sensor": 0.907130, "Linkage": 0.838551},
    "after": {"Sensor": -1.015798, "Linkage": 0.838551}
  },
  "review_events": [
    {"seq": 1, "actor_id": "demo-steward", "action": "reject",
     "argument_id": "S3", "reason": "B1 excludes serial suffix 2407"},
    {"seq": 2, "actor_id": "demo-steward", "action": "add",
     "argument_id": "S6", "evidence_ref": "N1",
     "reason": "Sensor replacement did not resolve the fault"},
    {"seq": 3, "actor_id": "demo-steward", "action": "authorize",
     "reason": "Linkage tag matches reviewed evidence"}
  ],
  "reviewer_id": "demo-steward",
  "authorized": true,
  "selected_tag": "damper_linkage_failure",
  "precedent_status": "pending_gate"
}
\end{lstlisting}

\noindent\textbf{Argument document (\texttt{argument\_records}).}
\begin{lstlisting}
{
  "_id": "argument-DEMO-2407-S1",
  "session_id": "session-DEMO-2407",
  "argument_id": "S1",
  "candidate": "Sensor",
  "effect": "support",
  "agent_role": "claims_telemetry",
  "model_version": "demo-model-v1",
  "prompt_version": "demo-prompt-v1",
  "source_ref": {"id": "C1", "field": "fault_code"},
  "tau": 0.70,
  "f10_before": 0.682928,
  "f10_after": 0.556808,
  "status": "active"
}
\end{lstlisting}

The embedded review events summarize a separate sequenced, append-only event log. Each remaining argument would have its own record with a cited source, base score, final strengths, and review status. A rejected argument such as S3 would retain its pre-review strength and rejection event rather than disappear from the audit trail. These records are proposed, not output from a deployed warranty system. An evaluation on independently adjudicated synthetic cases would need to measure tag accuracy, citation and applicability validity, reviewer interventions, and consistency across repeated runs.

\section{Discussion}
\label{sec:discussion}

The two cases show why a reviewer needs to challenge individual arguments, not only the selected metadata value. In Use case 1, the point name suggests one class until the manual locates the sensor in the mixed-air plenum. In Use case 2, the bulletin title appears to support sensor drift until its serial range excludes the unit. Retaining the original proposals, cited sources, and steward corrections makes these changes of reasoning inspectable. A versioned glossary helps interpret local terms, but it cannot establish whether a source applies to a particular unit or revision.

Use case 1 demonstrates a choice among Brick class labels, not validation of a Brick instance graph. The reported margins reproduce the reference implementation's three-label scoring and should not be read as ontology-normalized scores. An ontology-aware deployment would collapse equivalent class labels before scoring, record the ontology version and canonical class, link the point instance to its equipment through \nolinkurl{brick:hasPoint} or \nolinkurl{brick:isPointOf}, and validate the resulting graph against that Brick version. The warranty failure-mode tag in Use case 2 is governed metadata outside the Brick point-classification example.

The examples also expose limits of the illustrative scoring method. The warranty calculation demonstrates that the stored graph can be rescored after review; it does not establish that the selected tag is more accurate than an alternative method. The signed-sum margin depends on which claims are represented as separate arguments. S1 and S2 both derive from C1, for example, so repeated or correlated evidence could increase a margin without adding independent support. A deployment would need to identify shared provenance and test decision stability under alternative argument graphs as well as alternative values of $\tau$ and $\alpha$.

An auditable record makes a decision reconstructable, but it does not make the decision correct. Stewards still need to check document scope, resolve conflicting evidence, and authorize commits. Evaluation should measure review time, rejected arguments, erroneous commits, and later reversals alongside classification accuracy. The point-classification reference implementation demonstrates part of the workflow; the cross-system warranty record and promotion gate remain proposed. Their effects on review workload and decision quality have not been measured.

\section{Conclusion and Future Research}

Agents can perform much of the classification, linking, and documentation work on which data governance depends. Governance programs still need control over what enters the catalog. Contestable governance attaches cited structured and unstructured evidence to each proposal, lets a steward challenge it before commit, and stores the result as an attributable record retrievable in one query. A promotion gate controls what becomes precedent, while triage reserves full specialist argumentation for uncertain cases. The synthetic point-classification case illustrates how a reviewer can correct a name-based inference and preserve the reason for the correction. The illustrative warranty case shows how specialist agents and a versioned glossary could expose conflicts among claim codes, telemetry, technician observations, and bulletin scope before a steward commits a failure-mode tag. Requiring a steward for every correction may delay resolution, even when an edit is low risk and reversible. A future risk-based path could automatically correct a typographical error in a point description without changing its class, provided the evidence is unambiguous and the system records the original value, rationale, and means to reverse the edit. The workflow described here still requires review before commit; the conditions and effects of such an automatic path need evaluation.

Future research should test the learning loop. Once verified decisions are written back through the promotion gate, an evaluation could measure whether precedent improves accuracy on later units of the same equipment series, how far an incorrect precedent spreads before demotion, and when a firmware revision makes an older precedent unreliable. Separate comparisons should test whether versioned glossary entries and specialist roles improve citation validity and repeated-run consistency over a shared, unstructured prompt.

Validating the scoring rule requires formal and empirical tests. Formal checks should assess boundedness, the direction of support and attack effects, the treatment of isolated and duplicated arguments, and convergence on cyclic graphs against published axioms and analyses \citep{amgoud2018weighted,potyka2019}. An empirical comparison should hold the evidence, candidate set, and adjudicated labels fixed while comparing the signed-sum logistic rule with DF-QuAD \citep{rago2016} and a specified modular alternative, documenting how each handles reciprocal cycles. A versioned rubric should define how assessors assign $\tau$ from applicable evidence and test agreement among assessors. The rubric and $\alpha$ should be selected on development cases, then tested on held-out cases for top-class accuracy and ranking stability under small score changes, missing or duplicated arguments, and repeated agent runs. An error-bounded stopping rule should replace the fixed ten iterations so that each graph is scored to a stated numerical tolerance.

\bibliographystyle{plainnat}
\bibliography{references}

\end{document}